\documentclass[sigconf]{acmart}

\usepackage{booktabs} 
\newcommand{\myfirstpara}[1]{\noindent {\bf #1:}}
\newcommand{\mypara}[1]{\vspace{0.5em} \myfirstpara{#1}}

\setcopyright{rightsretained}
\usepackage{hyperref}
\usepackage[capitalize]{cleveref}
\crefname{section}{Sec.}{Secs.}
\Crefname{section}{Section}{Sections}
\Crefname{table}{Table}{Tables}
\crefname{table}{Tab.}{Tabs.}

\acmDOI{10.475/123_4}

\acmISBN{123-4567-24-567/08/06}

\acmConference[ICVGIP'24]{Indian Conference on Computer Vision, Graphics and Image Processing}{December 2024}{Bangalore, India}
\acmYear{2024}
\copyrightyear{2024}

\begin{document}
\title{Exploiting Contextual Uncertainty of Visual Data for Efficient Training of Deep Models}

\author{Sharat Agarwal \\ \href{https://sharat29ag.github.io}{\color{red}sharat29ag.github.io}}
\affiliation{%
 \institution{Indraprastha Institute of Information Technology, Delhi}
  \country{India}
}

\renewcommand{\shortauthors}{}

\begin{abstract}
Objects, in the real world, rarely occur in isolation and exhibit typical arrangements governed by their independent utility, and their expected interaction with humans and other objects in the context. For example, a chair is expected near a table, and a computer is expected on top. Humans use this spatial context and relative placement as an important cue for visual recognition in case of ambiguities. Similar to human's, DNN's exploit contextual information from data to learn representations. Our research focuses on harnessing the contextual aspects of visual data to optimize data annotation and enhance the training of deep networks. Our contributions can be summarized as follows: \texttt{(1)} We introduce the notion of contextual diversity for active learning CDAL \cite{cdal} and show its applicability in three different visual tasks semantic segmentation, object detection and image classification, \texttt{(2)} We propose a data repair algorithm \cite{wacv22} to curate contextually fair data to reduce model bias, enabling the model to detect objects out of their obvious context, \texttt{(3)} We propose Class-based annotation \cite{wacv23}, where contextually relevant classes are selected that are complementary for model training under domain shift. Understanding the importance of well-curated data, we also emphasize the necessity of involving humans in the loop to achieve accurate annotations and to develop novel interaction strategies that allow humans to serve as fact-checkers. In line with this we are working on developing image retrieval system for wildlife camera trap images and reliable warning system for poor quality rural roads. For large-scale annotation, we are employing a strategic combination of human expertise and zero-shot models, while also integrating human input at various stages for continuous feedback.
\end{abstract}


%
%



\maketitle

\section{Introduction} 

\textit{``For me context is the key - from that comes the understanding of everything.'' - Kenneth Noland}

Objects in the actual world exhibit typical arrangements, giving information of their interaction with other objects and the context of the overall scene. Humans use this spatial context as an essential cue for visual recognition in natural settings \cite{biederman1982scene}. When we look at a complex scene, we perceive it effortlessly and identify the objects even without recognizing them. Context of the objects in the real world helps us solve perceptual inference faster and more accurately. It’s generally observed that objects appearing in a consistent or familiar background are detected more accurately than objects in an inconsistent environment. For instance, we perceive the round disc-like object on the dining table from a distance as a diner plate.

Spatial context is an essential factor in facilitating scene understanding and object recognition for both machines and humans. Similar to how humans learn concepts and objects by observing their surroundings, deep learning models leverage large amounts of data, often annotated with labels, to learn diverse representations that can generalize to unseen data. One of the reasons why deep neural networks have done exceptionally well in the past decade is the availability of large diverse datasets. While data is crucial for deep learning, it is often overlooked in a learning setting. The more quality data a model is exposed to, the better it can capture intricate patterns, relationships, and nuances, leading to higher accuracy and robustness. While this works in theory, the sheer scale of data for practical applications comes at a high labor cost to label, especially in very specialized fields like medical or autonomous driving domains where the cost of running simulations to produce ground truth is very expensive.

Possible algorithmic solutions like active learning comes to let the algorithm iteratively pick most \emph{informative} data examples to be labeled from unlabeled datasets in a manner such that it is representative of the underlying data distribution to a near-optimal learner \cite{settles2012active}. Traditional AL techniques \cite{lewis1994heterogeneous,lewis1994sequential,joshi2009multi} have mostly been based on \emph{uncertainty} and have exploited the ambiguity in the predicted output of a model. Existing approaches that leverage these cues are still insufficient in adequately capturing the spatial and semantic context within an image and across the selected set.
%

Considering the critical role that data plays in model training, \textit{``we argue that along with the quantity of data, the quality of data needs attention.''} To this end, we propose that model training should be designed so that models are trained using \emph{contextually} diverse data to ensure they are accurate and unbiased while being efficiently trained. We thus investigate different aspects of contextual information from the available data and need of human in the loop for efficient annotation.

\section{Progress and Results} \label{sec:section2}
As motivated in previous section contextual information is a crucial part for humans visual understanding and inspired deep networks. In this section, we focus on the main contribution of this manuscript, discussing the importance of contextual information in selecting data for visual recognition in different applications. 

\mypara{Contextual Diversity for Active Learning \cite{cdal},\texttt{ECCV20}}  
State of the art Active Learning approaches typically rely on measures of visual diversity or prediction uncertainty, which are unable to effectively capture the variations in spatial context. On the other hand, modern CNN architectures make heavy use of spatial context for achieving highly accurate predictions. Since the context is difficult to evaluate in the absence of ground-truth labels, we introduce the notion of \emph{contextual diversity} that captures the confusion associated with spatially co-occurring classes. Contextual Diversity (CD) hinges on a crucial observation that the probability vector predicted by a CNN for a region of interest typically contains information from a larger receptive field. Such a measure would help select a training set that is diverse enough to cover a \emph{variety of object classes} and their \emph{spatial co-occurrence} and thus improve generalization of CNNs. The objective of this paper was to achieve this goal by designing a novel measure for active learning which helps select frames having objects in diverse contexts and background.


\mypara{Contextually Fair Data To Reduce Model Bias \cite{wacv22}, \texttt{WACV22}} 
Co-occurrence bias in the training dataset may hamper a DNN model's generalizability to unseen scenarios in the real world. For example, in COCO\cite{lin2014microsoft} dataset, many object categories have a much higher co-occurrence with men compared to women, which can bias a DNN's prediction in favor of men. Recent works have focused on task specific training strategies to handle bias in such scenarios, but fixing the available data is often ignored. We propose a novel and more generic solution to address the contextual bias in the datasets by selecting a subset of the samples, which is fair in terms of the co-occurrence with various classes for a protected attribute. We introduce a data repair algorithm using the coefficient of variation, which can curate fair and contextually balanced data for a protected class(es). This helps in training a fair model irrespective of the task, architecture or training methodology. Proposed solution is simple, effective, and can even be used in an active learning setting where the data labels are not present or being generated incrementally. 

\mypara{Contextual Class for Active Domain Adaptation \cite{wacv23}, \texttt{WACV23}} 
In Active Domain Adaptation (ADA), one uses Active Learning (AL) to select a subset of images from the target domain, which are then annotated and used for supervised domain adaptation (DA). Given the large performance gap between supervised and unsupervised DA techniques, ADA allows for an excellent trade-off between annotation cost and performance.  Prior art makes use of measures of uncertainty or disagreement of models to identify \emph{regions} to be annotated by the human oracle. However, these regions frequently comprise of pixels at object boundaries which are hard and tedious to annotate. Hence, even if the fraction of image pixels annotated reduces, the overall annotation time and the resulting cost still remain high. In this work, we propose an ADA strategy, which given a frame, identifies a set of classes that are hardest for the model to predict accurately, thereby recommending semantically meaningful regions to be annotated in a selected frame. We show that these set of \emph{hard} classes are context-dependent and typically vary across frames, and when annotated help the model generalize better. We propose two ADA techniques: the \texttt{Anchor-based} and \texttt{Augmentation-based} approaches to select complementary and diverse regions in the context of the current training set.

\section{Future Plan}
Work done conducted so far investigates the various aspects of contextual information within the available data for annotation and effectively training DNN's. But it is important to emphasize the necessity of involving human-in-the-loop to achieve accurate annotations and to aid as a fact-checker. Human-in-the-loop approaches typically rely on humans completing the task independently and then combining their opinions with an AI model in various ways, as these models offer very little interpretability or correctability. Human-in-the-loop opportunity is to look to novel interaction strategies to break this away from traditional pipeline. Instead, we should focus on designing systems that allow for flexible human intervention at different stages of the AI process, thereby optimizing both human time and effort. Prior work in health AI \cite{peiffer2020machine} has sought to achieve better accuracy via decision-support tools for clinicians that assist humans in decision-making. Human-in-the-loop participatory systems have also been proposed to accurately and robustly categorize wildlife images \cite{bondi2022role}. 

In line with these proposed works, we are also working on developing a image retrieval system for wildlife camera trap images and reliable warning system for poor quality rural roads. Developing both the systems requires \texttt{(1)} annotating large amount of unlabeled diverse data which we are approaching through a smart combination of human expertise and zero-shot models as a few minutes of annotation by users can lead to sizeable gains over these zero-shot classifiers and \texttt{(2)} involving human expertise at different levels for feedback and correctness. Developing these tools require a user-centric approach to developing real-world classifiers for these subjective concepts. As end users are usually not machine learning experts thus we need interactive systems with human feedback that elicit the subjective decision boundary from the user. It would be very interesting to explore how a cyclical feedback loop between the oracle and the model improves both of them. 

\section{Conclusion}
This paper investigates the contextual aspect of visual data and use it for training deep networks effectively. We proposed techniques to efficiently annotate and use data for different computer vision applications, reducing cost for annotation and training. Further we plan to extend our work in developing novel interactive human-in-the-loop systems where human's can intervene at different level with their expertise. We plan to compile outputs from our research such that it is helpful in different interdisiplinary areas. Where we can help and use our knowledge of understanding deep models and data requirements to make real-world systems more \texttt{reliable}, \texttt{efficient} and \texttt{trustworthy}. 


\bibliographystyle{ACM-Reference-Format}
\bibliography{ICVGIP-Latex-Template}

\end{document}